\documentclass[conference]{IEEEtran}
\IEEEoverridecommandlockouts
\usepackage{cite}
\usepackage{amsmath,amssymb,amsfonts}
\usepackage{algorithmic}
\usepackage{graphicx}
\usepackage{textcomp}
\usepackage{xcolor}
\def\BibTeX{{\rm B\kern-.05em{\sc i\kern-.025em b}\kern-.08em
    T\kern-.1667em\lower.7ex\hbox{E}\kern-.125emX}}
\newcommand{\myreferences}{Bibl.bib}

\usepackage{multirow}
\usepackage{subcaption,graphicx}
\usepackage[hidelinks]{hyperref}
\begin{document}

\title{Exploiting Individual Graph Structures to Enhance Ecological Momentary Assessment (EMA) Forecasting \\
\thanks{This study is part of the project ‘New Science of Mental Disorders’ (www.nsmd.eu), supported by the Dutch Research Council and the Dutch Ministry of Education, Culture and Science (NWO gravitation grant number 024.004.016).
}
}

\author{\IEEEauthorblockN{Mandani Ntekouli}
\IEEEauthorblockA{\textit{Department of Advanced Computing Sciences} \\
\textit{Maastricht University}\\
Maastricht, The Netherlands \\
m.ntekouli@maastrichtuniversity.nl}
\and
\IEEEauthorblockN{Gerasimos Spanakis}
\IEEEauthorblockA{\textit{Department of Advanced Computing Sciences} \\
\textit{Maastricht University}\\
Maastricht, The Netherlands\\
jerry.spanakis@maastrichtuniversity.nl}
\and
\IEEEauthorblockN{Lourens Waldorp}
\IEEEauthorblockA{\textit{Department of Psychological Methods} \\
\textit{University of Amsterdam}\\
Amsterdam, The Netherlands\\
L.J.Waldorp@uva.nl}
\and
\IEEEauthorblockN{Anne Roefs}
\IEEEauthorblockA{\textit{Faculty of Psychology and Neuroscience} \\
\textit{Maastricht University}\\
Maastricht, The Netherlands\\
a.roefs@maastrichtuniversity.nl}
}
\maketitle

\begin{abstract}
In the evolving field of psychopathology, the accurate assessment and forecasting of data derived from Ecological Momentary Assessment (EMA) is crucial. EMA offers contextually-rich psychopathological measurements over time, that practically lead to Multivariate Time Series (MTS) data. Thus, many challenges arise in analysis from the temporal complexities inherent in emotional, behavioral, and contextual EMA data as well as their inter-dependencies. To address both of these aspects, this research investigates the performance of Recurrent and Temporal Graph Neural Networks (GNNs). Overall, GNNs, by incorporating additional information from graphs reflecting the inner relationships between the variables, notably enhance the results by decreasing the Mean Squared Error (MSE) to $0.84$ compared to the baseline LSTM model at $1.02$. Therefore, the effect of constructing graphs with different characteristics on GNN performance is also explored. Additionally, GNN-learned graphs, which are dynamically refined during the training process, were evaluated. Using such graphs showed a similarly good performance. Thus, graph learning proved also promising for other GNN methods, potentially refining the pre-defined graphs.

\end{abstract}


\section{Introduction}
In the field of psychopathology, recent advancements in Ecological Momentary Assessment (EMA) for real-time monitoring have offered new opportunities to uncover the complex world of mental disorders \cite{becker2016predict, torous2018smartphones}. EMA provides a tool to capture the individual dynamics over various factors, regarding emotions, behaviors, and physiological states in their daily life \cite{fried2017moving}. All these can be measured multiple times throughout the day over a period of days, weeks, or even months. Such data, structured as Multivariate Time-series (MTS), hold a significant amount of information to better understand individuals' underlying mechanisms and consequently study mental disorders. Particularly, EMA data could be used for accurately predicting and forecasting the progression of variables related to the progression of psychopathological conditions \cite{ntekouli2022using}. The ability to understand and predict such conditions is crucial for the development of early intervention and tailored treatment strategies \cite{roefs2022new}.

The paradigm of forecasting in multivariate time series presents a unique confluence of challenges and opportunities in the domain of Machine learning (ML) \cite{lim2021time, tealab2018time}. Such challenges are increasingly complex due to the inherent temporal dynamics, the interactions across time-series, as well as the high dimensionality of the data.

While traditional forecasting methods, such as the Vector Autoregressive (VAR) model and Autoregressive Integrated Moving Average (ARIMA), focus on handling the temporal aspect of the data, they often fall short when high dimensional and interdependent variables are involved, as in EMA datasets \cite{wild2010graphical, fried2017moving}. These limitations highlight the need for more sophisticated approaches to handle the complex nature of EMA multivariate time-series. Therefore, deep learning methodologies have been established as a milestone of multivariate time-series analysis \cite{lim2021time, tealab2018time}. More specifically, Deep Learning (DL) methodologies designed for time-series, such as Recurrent Neural Networks (RNN) models, have been effective due to their ability to model complex non-linear relationships on the temporal aspect of the data as well as handling large volumes of data. Although RNN can successfully identify short-term dependencies among data, global or long-term effects are not taken into account \cite{rnn}.
Moreover, these models do not address the spatial aspect of data, adding another layer of complexity. Thus, these challenges call for an innovative approach that can adeptly handle both the long-term temporal and spatial domains inherent in EMA MTS data. 

Following the theory of Complex Networks, the time series dynamics can be captured not just by data sequences, but as a complex network of interactions, offering a multidimensional perspective of the data \cite{gao2017complex}. Networks, represented by graphs, are a powerful mechanism, able to characterize the relationships between all variables. Therefore, an additional transformation from MTS to a network representation is promising as an alternative way for time-series analysis.

Inspired by this, graph-based neural networks (GNNs) have emerged as promising models to address all the challenges mentioned above \cite{wu2020comprehensive}. These methods exploit the inherent structure in data to capture complex dependencies, offering potentially higher accuracy by incorporating a graph representing variables’ interactions. Therefore, a well-defined and reliable graph representation is the key to enhance the efficacy of the applied forecasting model.

In this paper, the use of GNN models is explored in the context of processing EMA data. To the best of our knowledge, this is the first application of GNNs on EMA with a goal to enhance the prediction of future variables related to psychopathology. Also, it is investigated how different graph structures, representing different information about variables’ interactions, impact the GNN performance. Through this, the nature of each graph is explored. 
A graph leading to a better model performance potentially holds a better representation of the data's complex patterns and relationships. As an additional contribution, the exploration goes beyond only utilizing pre-defined graphs but also graph learning approaches. Graphs, adaptively updated during the training process, could potentially provide more insights into hidden inter-variables connections. Thus, it is believed that GNNs and graph learning techniques set new methodological directions for understanding and forecasting complex psychopathological trends.

\section{Related Work}
Conducting a literature review on the relevant EMA forecasting approaches is crucial to fully understand the scope of this study. First, the existing research on EMA data is reviewed. 
Starting from linear models to more advanced models focusing on the temporal aspects of the data, such as Recurrent Neural Networks (RNNs), we finally move to explore the GNN models, capable of handling the whole spatio-temporal spectrum of the data.

\subsection{Forecasting on EMA Data}
In the field of psychopathology, the recent paradigm shift to the network approach brought changes in the way of studying mental disorders \cite{borsboom2017network}. The network approach conceptualizes mental disorders as complex systems comprising of dynamically interacting symptoms, behaviors, and contextual factors \cite{fried2017moving}. In practice, network modeling was studied in the form of EMA time-series forecasting task. Most of the studies then focus on applying linear statistical models, like the VAR model, with an aim to capture the inter-connections among multiple time series variables \cite{wild2010graphical, epskamp2018gaussian}. Exploiting the inherent interpretation of linear models offers an insightful way to capture disorders, represented as networks or graphs of all co-occurring variables. Nevertheless, there are significant challenges arise by using VAR models in reliably forecasting short-term psychopathological fluctuations. More specifically, the complex non-stationary nature of EMA data cannot accurately be reflected by the produced linear interactions, while high data dimensionality involving interdependent variables can yield unstable results.

In a way to incorporate non-linear interactions, a switch to Machine Learning and Deep Learning models was undertaken.
More advanced ML models were studied in \cite{ntekouli2022using, ml}, while models particularly focusing on the temporal aspect of EMA, such as RNN, were introduced in \cite{koppe2019recurrent}. In general, RNN-based models have been widely applied in the task of time-series forecasting due to their ability to handle large volumes of data and capture complex non-linear temporal relationships. Despite their success in identifying local relationships, global effects or spatial interactions are not taken into account. Thus, these challenges add another layer of complexity, calling for an innovative approach that can address both the long-term temporal and spatial domains inherent in EMA MTS data. 

\subsection{Application of GNN on MTS}
Building on the VAR model and the principles of Complex Networks theory, representing variable interactions in networks to understand the whole system, it was believed that MTS forecasting can be naturally approached from a graph perspective. Thus, modeling MTS data using GNNs seems as a promising strategy. GNNs could preserve the temporal nature of data while exploiting the inter-dependency among variables of time series. The key component of GNNs is the use of an externally defined, well-structured graph. However, ground truth knowledge about graph structure is not always available in the context of MTS applications. Based on this, GNNs are categorized into prior-knowledge based or learning-based models. According to the first category, models exploit some additional information available. For example, graphs can be constructed by the physical connections existing in many scenarios such as road traffic 
\cite{a3tgcn, astgcn}. Alternatively, a graph adjacency matrix can be created by capturing connectivity in terms of similarity or distance \cite{measures,simTSC}. On the other hand, graph-learning methods automatically update the graph matrix while optimizing the training loss. This could potentially facilitate uncovering hidden inter-relationships among variables. Examples of works incorporating graph-learning methods are given in \cite{mtgnn, nri, gts, graph_cat}. 

\section{Methodology}
In this section, the methodology to approach the MTS forecasting problem through GNNs is discussed in detail. All steps of the proposed methodology are presented in Fig.~\ref{fig:forecast}. First, the nature of EMA MTS data is provided along with the problem formulation of the current forecasting task. Afterwards, we continue by demonstrating the workflow of applying GNN models adapted to the application of EMA data. As already clarified, the important role of graph construction is also explored, where static distance metrics as well as GNN-learned are evaluated. Therefore, all the key components are separately described as follows:

\subsection{EMA Data}
As already introduced, EMA offers an advanced way to gather real-time data about individuals' psychopathology-related behaviors, emotions, and symptoms. According to the EMA protocol for data collection, participants respond to digital questionnaires on their smartphones, where they rate the perceived intensity of different EMA items (including positive and negative affect, stress, impulsivity, etc.) along with contextual information. Ratings are made on a 7-point Likert scale, which, after being normalized for each individual, allows the data to be analyzed as continuous data.

In practice, as data collection occurs multiple times on a daily basis, EMA is typically structured as MTS data. To elaborate, we denote the entire EMA dataset as $X$, including the individual EMA data of the total number of participants, $N$. The $X$ dataset can be represented as $X=\{X_1, X_2,..,X_N\}$. On the individual level, data $X_i$ are comprised of $V$ variables, each recorded over several time-points, $T_i$. Although the total number of time-points may differ among participants, because of possible missed/unanswered questionnaires, for each individual $i$, all $V$ variables are collected over the same $T_i$ period. Based on these characteristics, every individual EMA dataset $X_i$ represents a multivariate time-series dataset, structured as $\{X_{i, 1..V,1..T_i}\}$.

\subsection{Forecasting Problem}

\begin{figure}
    \centering
    \includegraphics[width = 0.9\columnwidth]{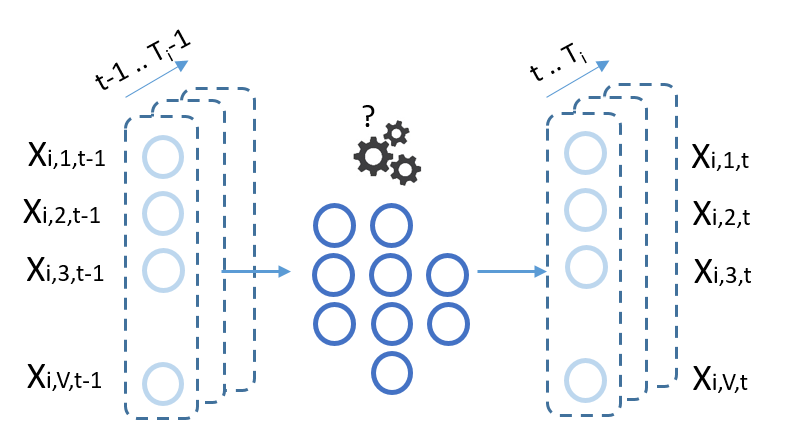}\caption{Personalized forecasting models, repeated across $N$ individuals, aiming at accurately predicting the 1-lag future responses of all $V$ variables.}
    \label{fig:forecast}
\end{figure}

In the field of psychopathology, a critical task is using the EMA data to accurately forecast the future of individual psychological variables. More specifically, in this study, the goal is to predict the 1-lag future values for all variables. That is, for each $X_i$, using as input all variables at time-point $t-1$, $X_{i,1..V,t-1}$, the goal is to predict all variables at time-point $t$, $X_{i,1..V,t}$. In practice, since the output is set as the whole range of variables at the time-point $t$, the number of output states is equal to the input states, that is $V$. This individual setup is also described in Fig.~\ref{fig:forecast}. Considering the temporal nature of data, the temporally-oriented NNs are capable of processing data inputs of varying sequences. Therefore, the dimensions of the input data could differ, by using a single-step or multi-step sequences for all the variables.

\subsection{GNN Models}
Graph Neural Networks (GNNs) have emerged as a powerful framework for learning representations on graph-structured data. They extend traditional NNs to handle graph data by propagating information (message passing) through the graph’s nodes and edges, thereby capturing the complex relationships and structures inherent in different types of data. Focusing on MTS applications, the key to their success is the ability to model both spatial and temporal dependencies within the data, which is particularly valuable in dynamic variables whose relationships evolve over time.

GNN models can be generally categorized based on the primary processing mechanisms they use. The three prominent categories include Recurrent Graph Convolution (R-GCN), Temporal Graph Attention (T-GAT), and Auxiliary Graph Convolution models \cite{wu2020comprehensive,rozemberczki2021pytorch}. R-CGNs, based on RNN principles,
capture the temporal dependencies by maintaining a hidden state that evolves over time. By incorporating such architectures, they can effectively propagate information across the graph structure over multiple time steps, allowing the model to learn complex spatial-temporal patterns. 

On the other hand, T-GAT models make use of attention mechanisms to weigh the importance of different time steps dynamically. This approach allows the model to focus on the most relevant parts of the graph at each time step, which is beneficial for tasks where the significance of different nodes can vary significantly over time. Naturally, the Auxiliary models collect elements from both recurrent and attention categories to capture a richer set of spatial-temporal patterns.

For the following analysis, we study a subset of GNNs belonging to these categories, specifically chosen because of their ability to be trained on multi-step sequential data. 
First, A3TGCN (implementation of the Attention Temporal Graph Convolutional Networks), from the R-GCN category, stands out for its integration of Graph
Convolutional Networks (GCN) with Gated Recurrent Units (GRU), offering a
robust framework for processing lower-order proximity data \cite{a3tgcn}. Next, ASTGCN (Attention Based Spatial-Temporal Graph Convolutional Networks) and
MTGNN (Multivariate Time Series Forecasting Graph Neural Networks), both belonging to the T-GATs category, are investigated \cite{astgcn, mtgnn}. While ASTGCN employs an attention mechanism
coupled with Chebyshev convolutions, MTGNN incorporates custom
attention mechanisms, offering advanced approaches to higher-order proximity data
analysis. Moreover, the latter model incorporates a graph-learning module.

\subsection{Graph Construction}
As already clarified, the effectiveness of GNNs is expected to be influenced by the structure of the graph and the information it incorporates. Given this impact, it becomes essential to explore and evaluate various graph construction strategies. First, 
measures to count the similarity between variables within the graph are employed \cite{measures, trajectory,simTSC}. Such graphs could uncover the inter-dependencies among EMA variables, providing a structured way to quantify and visualize these. It is also crucial to acknowledge that because of individual variability, different patterns of variables' inter-relationships would be identified for each one. Thus, integrating and learning from the structure of individual-specific similarity-based graphs could eventually enhance the personalized GNN models.

The exploration of similarity-based graphs starts with the utilization of a classic distance metric, the Euclidean distance. An extension of this is the k-Nearest Neighbors (kNN) approach, which incorporates only the information of significant edges, and particularly the Euclidean distance of $k$ connections per node \cite{measures}. Additionally, two more temporally-oriented metrics are involved. These are the Dynamic Time Warping (DTW) as well as the Pearson correlation. DTW is particularly useful in measuring similarities by aligning sequences that may vary in time or speed. This is a crucial aspect because it is common for emotions and psychopathology symptoms to fluctuate differently over time. For instance, the effect of an event on each variable as well as their response time are not expected to be temporally synchronized, necessitating the need for proper alignment. On the other hand, Pearson correlation offers insights into the linear relationship between variables over time. Thus, these measures provide a valuable assessment of different variable similarities, crucial for optimizing GNN performance.

Beyond only utilizing graph similarity measures, another critical aspect of enhancing GNN performance involves exploring the graphs learned by some special GNN models that integrate a graph learning module. During training, the way these GNNs interpret, utilize, and update the connections within graph adjacency matrices could improve their overall graph effectiveness. Therefore, a better performance could be caused by more informative learned graph structures. By such graphs, we can get invaluable insights into how GNNs process the complex relationships between variables highlighting the underlying mechanisms and patterns that GNNs identify during their learning process. Subsequently, enhanced graph information could be also used as an input to other graph-based models and potentially boost their performance.

\begin{figure*}
    \centering
\includegraphics[width = 0.6\textwidth]{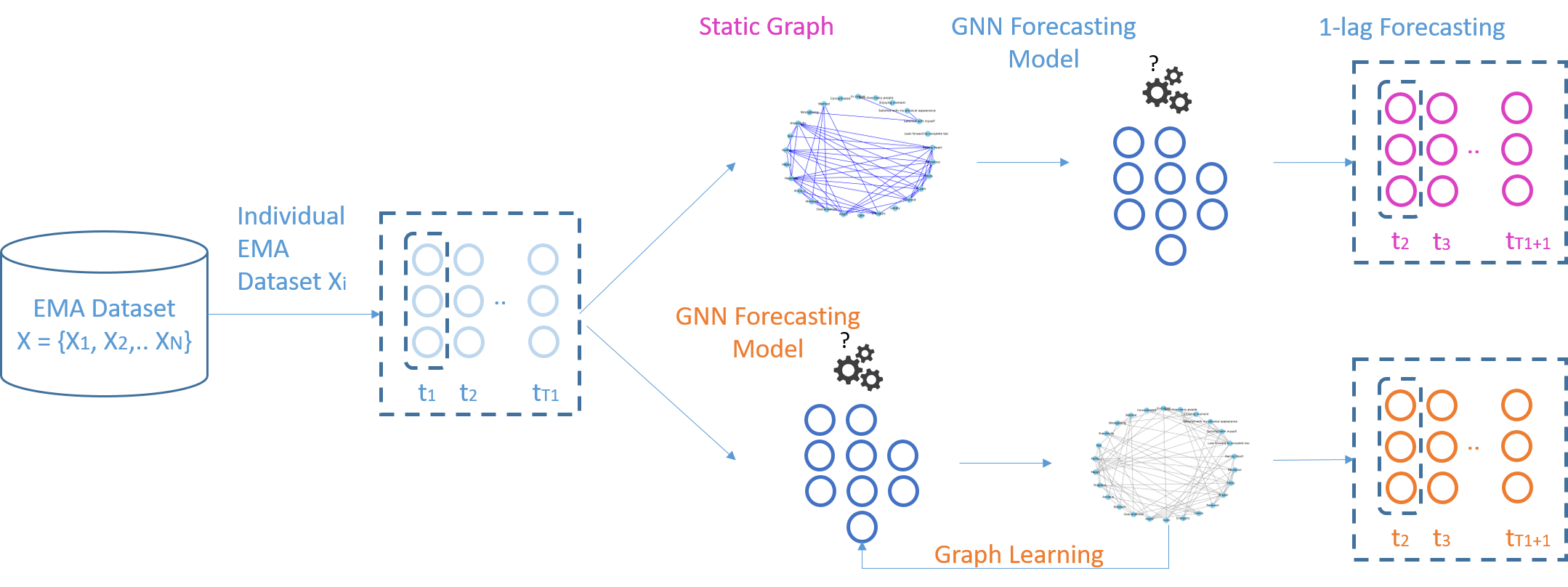}
\caption{Proposed experimental framework for investigating the effect of different static graphs and graph learning on the forecasting performance of various GNN models.}
\label{fig:gnn_method}
\end{figure*}

\section{EMA Dataset}
The dataset examined consists of real-world EMA pilot data, collected sequentially over
time across students in Dutch Universities \cite{roefs2022new, martinez2023developing}. Particularly, a group of 269 participants was asked to
complete 8 questionnaires per day at fixed time intervals throughout 28
days. The questionnaires included a series of questions regarding
their experiences, emotions, and behaviors. Subsequently, their responses, recorded on a 7-point Likert scale, could be used to provide insights into each participant's psychological state.

During pre-processing, each individual was analyzed separately. First, individuals with low compliance were eliminated, ensuring that the dataset consisted of active participants with the largest possible number of observations. 
Then, regarding the EMA variables, a number of them with low variance were also removed. Finally, a group of 100 individuals remained for analysis. After examining the variables kept for each individual, they were all eventually represented by the same subset of 26 variables. These selected variables spanned across an average of 140 time-points. 

\section{Experimental Setup}
\label{setup}
The comparative analysis focuses on three main aspects. First, the comparison of models' performance, that is examining the extent to which GNN models can be effective after incorporating graph information. Second, the impact of the nature of connectivity on the constructed graph, representing the inner relationship between the variables. Finally, the effectiveness of the graph learning module compared to pre-defined graphs. All the examined parameters regarding GNN models, graph structure and graph sparsity are presented in Table \ref{tab:param}.

\begin{table}[t]
    \centering
    \caption{All examined scenarios regarding GNN models, graph structure and graph sparsity}
    \begin{tabular}{c|c|c}
        \textbf{GNN Models} & \textbf{Graph Structure} & \textbf{Graph Sparsity} \\
        \hline
        \hline
        A3TGCN & Euclidean & $20\%$\\
        ASTGCN & kNN & $40\%$\\
        MTGNN & DTW & $100\%$ \\
        & Correlation &\\
        & GNN-learned &\\
        & Random &\\
        \hline
    \end{tabular}
    \label{tab:param}
\end{table}

\subsection{Experiment A: Investigating Different GNNs}
In Experiment A, a comparative analysis of the performance across various GNN models is conducted. The whole set of three models, A3TGCN, ASTGCN and MTGNN, is trained and evaluated on the last part of the data across all individuals. Then, all GNN models are assessed against the widely-applied Long short-term memory (LSTM). LSTM is commonly used for multivariate
time series forecasting based on its ability to capture complex relationships
within sequential data.
It is then important to determine whether GNN is suitable for forecasting EMA data and to what extent these can outperform the baseline model. Thus, Experiment A will provide a clearer understanding of the effectiveness of GNNs in handling EMA data
complexities compared to the LSTM model.

\subsection{Experiment B: Investigating the Efficacy of Different Graph Structures}
Since graph information is a key factor for a successful GNN model, different graph construction techniques are investigated. The primary objective is to determine which
distance metric is the most suitable when the underlying graph
topology is not predefined, a scenario frequently encountered in EMA datasets. Therefore, the performance of GNNs using 4 different metrics, such as Euclidean distance (EUC), kNN, DTW and cross-correlation (CORR) is evaluated \cite{measures}. For an additional check, the performance is compared also to cases when a random graph is used as an input, meaning that no useful information is involved. 

As another aspect of graph construction, the level of sparsity in connectivity is assessed. That is, 
investigating the effect of employing sparser graphs,
with fewer connections among variables, compared to more densely connected
graphs. Although it is generally known that sparse graphs are more effective for training GNNs \cite{mtgnn}, this experiment involves three distinct graph density thresholds (GDT), keeping high sparsity with $20\%$ and $40\%$ of the graph edges or no sparsity with $100\%$. By comparing the outcomes from graphs with varying GDT, it can be determined whether a more interconnected
network of variables or a more selective one could enhance the forecasting accuracy and model efficiency.

\subsection{Experiment C: Investigating Static and GNN-Learned Graphs}

The core objective of this experiment is to assess the GNN performance after using a static or GNN-learned graph structure. More specifically, the impact of the 4 static distance-based graphs, as above analyzed, is compared to the graph learning procedure applied through MTGNN. During training, the MTGNN model continuously
updates its graph topology while starting from an initial graph structure or a random one. Apart from only assessing MTGNN using its learned graph, the extracted learned representations could be used as an input to the rest of the GNN models. This would facilitate determining whether a dynamically-learned graph by MTGNN could be more informative than distance-based graphs. In the case that the MTGNN-learned graph could also enhance the predictive performance of ASTGCN and A3TGCN, then valuable insights could be uncovered by that graph's information.

\subsection{Model Parameters}
In this study, all models investigated were developed utilizing the PyTorch Temporal Geometric framework \cite{rozemberczki2021pytorch}. However, to take full advantage of these models, the optimal setting of hyperparameters needs to be determined. The process of fine-tuning these parameters is essential for model effectiveness.

First, during the training phase, a core set of hyperparameters was evaluated for each model, including the learning rate, batch size, and the number of training epochs. After exploring various choices, the Adam algorithm is used as the optimizer with a learning rate of 0.01. Since our modeling is designed in a personalized approach, each individual's data is processed in a single batch, and training is iterated over 300 epochs. To avoid overfitting, we adopted a dropout strategy with a rate of 0.3.

Furthermore, another critical group of hyperparameters was examined regarding the number of hidden units across all possible channels or layers within each model. For example, in MTGNN, convolution, temporal and skip-connection layers are involved. Because of the relatively short-length sequences in each EMA dataset, there was no need to incorporate a very deep network or very large filters. Through experiments using 16 or 32 hidden units, we determined that setting all layers to 32 hidden units, along with a kernel size of $k=3$, yielded the optimal performance. For other model-specific hyperparameters, their default settings were retained.

\subsection{Models Evaluation}
Performance is measured in terms of Mean Squared Error (MSE), averaged across all $N$ individuals. For each individual data $X_i$ then, the errors are assessed on the test set (where time-points range from 1 to $T$), while predicting all $V$ variables. Since each $X_i$ is time-series data, these are sequentially split into training (first $70\%$ of each dataset $X_i$) and test (the last $30\%$ of each dataset). Particularly, the calculation of the MSE error between the true $X$ and the predicted $\hat{X}$ values is described in \eqref{eq:loss}.

\begin{equation}
\label{eq:loss}
    L_{MSE} = \frac{ \sum_{i=1}^N \sum_{t=1}^{T} \sum_{v=1}^{V} (X_{i,v,t} - \hat{X}_{i,v,t})^2}{N\cdot T\cdot V}
\end{equation}


\section{Experimental Results}
The analysis is designed to assess the effectiveness of GNNs in handling EMA data through a series of three structured experiments, as described in Section \ref{setup}. 

\subsection{Experiment A}
Experiment A investigates the comparison results between GNN models with the baseline LSTM model as well as across all GNN models. Table \ref{tab:ExpA} shows the results of the three examined GNN models based on the four static graphs
(EUC, DTW, kNN, and CORR), while keeping sparsity at
$GDT = 20\%$. Also, both single- (Seq1) and multi-step (Seq2, Seq5) inputs are given. The presented scores refer to the MSE values averaged across all individuals.

\begin{table}[ht!]

\caption{Comparison of GNN models with LSTM when using single- and multi-step input. The best scores are highlighted.}
\label{tab:ExpA}
\centering
\begin{tabular}{|l|c|c|c|}
\hline
Model &     Seq1 &     Seq2 &     Seq5 \\
\hline
\hline
Baseline LSTM & 1.027(0.492) & 1.020(0.484)& 1.022(0.499)\\
\hline
\hline
A3TGCN$_{EUC}$  &  1.032(0.500) &  1.034(0.493) &  1.034(0.506) \\
ASTGCN$_{EUC}$  &  0.907(0.488) &  0.881(0.459) &  0.885(0.442) \\
MTGNN$_{EUC}$   &  0.868(0.430) &  0.851(0.430) &  0.845(0.432) \\
\hline
\hline
A3TGCN$_{DTW}$  &  1.032(0.500) &  1.033(0.490) &  1.034(0.504) \\
ASTGCN$_{DTW}$  &  0.900(0.482) &  0.886(0.470) &  \textbf{0.883(0.442)} \\
MTGNN$_{DTW}$  &  0.871(0.430) &  0.855(0.427) &  0.846(0.430) \\
\hline
\hline
A3TGCN$_{kNN}$  &  1.032(0.500) &  1.037(0.502) &  1.035(0.505) \\
ASTGCN$_{kNN}$  &  0.911(0.486) &  0.888(0.460) &  0.893(0.447) \\
MTGNN$_{kNN}$   &  0.862(0.428) &  0.844(0.422) &  \textbf{0.841(0.430)} \\
\hline
\hline
A3TGCN$_{CORR}$ &  1.020(0.494) &  1.026(0.493) &  1.027(0.501) \\
ASTGCN$_{CORR}$ &  0.908(0.488) &  \textbf{0.882(0.449)} &  0.885(0.438) \\
MTGNN$_{CORR}$  &  0.860(0.428) &  0.842(0.426) &  \textbf{0.840(0.431)} \\
\hline
\end{tabular}
\end{table}


Table \ref{tab:ExpA}, demonstrates that the GNN models provide enhanced performance compared to baseline LSTM, in terms of the average MSE score. In more detail, the best outcome is found when using a multi-step input, achieving an average MSE score of $0.840$ for MTGNN$_{CORR}$ and MTGNN$_{kNN}$, and around $0.845$ for MTGNN$_{EUC}$ and MTGNN$_{DTW}$. These are
followed by the ASTGCN$_{DTW}$ model with $0.883$.
In contrast,
the baseline LSTM performs worse, with an average MSE at $1.022$ when using the 5-timestep sequence. It is interesting to observe that only A3TGCN provides similar or slightly higher than LSTM errors, with MSE at $1.03$. This is reasonable because A3TGCN employs a quite simpler framework architecture, consisting of a temporal GCN along with an attention mechanism. In contrast, beyond MTGNN which incorporates graph learning, ASTGCN incorporates more complex blocks, such as spatial and temporal attention mechanisms, in addition to the fundamental GCN and convolution layers. This additional complexity in ASTGCN allows for better spatiotemporal representations within the data. 
Thus, apart from that case, GNN models outperform LSTM, noticing that MTGNN incorporating graph learning features achieved the best performance. 

Moreover, similar patterns in results are seen when using the single-step sequence of input data. GNN-models outperform LSTM, except for A3TGCN. Whereas, between the 1-step, 2-step, and 5-step input data, the results are also a bit improved in the latter case for all models. Therefore, models, using
a snapshot of the data at each moment, potentially missing valuable long-term patterns and fluctuations. 

\subsection{Experiment B}
In Experiment B, the assessment focuses on the impact of various graph construction techniques on GNN forecasting performance. The exploration was centered around
identifying the most
effective distance-based graph when the underlying graph topology is
not pre-defined. To verify the validity of our approach, these four examined metrics were also tested against a randomly
generated graph with the same amount of connected edges. Specifically, sparse graphs considering
$GDT = 20\%$ and $GDT = 40\%$ as well as dense with 
$GDT = 100\%$ were analyzed and their results are demonstrated in Table \ref{tab:gdt}.
The presented findings refer to 5-step input data training, since it was already observed that single- and multi-step follow the same trends in results. Moreover, regarding the random scenario, the average score after using 5 randomly generated in training is given.

\begin{table}[t!]
\caption{Average MSE errors for different levels of graph sparsity (GDT) when using multi-step input data. The best scores are highlighted.}
    \label{tab:gdt}
    \centering
    \begin{tabular}{|l|c|c|c|}
\hline
Model &     GDT = 20\% &     GDT = 40\% & GDT = 100\%\\
\hline
\hline
A3TGCN$_{EUC}$  &  1.034(0.506) &  1.035(0.505) &  1.032(0.496) \\
ASTGCN$_{EUC}$  &  0.885(0.442) &  0.878(0.430) &    0.879(0.432)  \\
MTGNN$_{EUC}$   &  0.845(0.432) &  0.843(0.430) &  0.843(0.430) \\
\hline
A3TGCN$_{DTW}$  &  1.034(0.504) &  1.035(0.505) &  1.034(0.503) \\
ASTGCN$_{DTW}$  &  0.883(0.442) &  0.883(0.434) &  \textbf{0.844(0.440)} \\
MTGNN$_{DTW}$   &  0.846(0.430) &  0.845(0.434) &  0.848(0.435) \\
\hline
A3TGCN$_{kNN}$  &  1.035(0.505) &  1.035(0.505) &    1.035(0.505)  \\
ASTGCN$_{kNN}$  &  0.893(0.447) &  0.875(0.429) &  0.882(0.443) \\
MTGNN$_{kNN}$   &  \textbf{0.841(0.430)} &  0.847(0.425) &  0.851(0.437) \\
\hline
A3TGCN$_{CORR}$ &  1.027(0.501) &  1.037(0.507) &  \textbf{0.970(0.385)} \\
ASTGCN$_{CORR}$ &  0.885(0.438) &  0.884(0.436) &  \textbf{0.840(0.436)} \\
MTGNN$_{CORR}$  &  0.840(0.431) &  0.841(0.434) &  \textbf{0.838(0.434)} \\
\hline
\hline
A3TGCN$_{RAND}$ &  1.032(0.504) &  1.033(0.506) &  1.033(0.504) \\
ASTGCN$_{RAND}$ &  1.059(0.622) &  1.062(0.633) &    1.062(0.630) \\
MTGNN$_{RAND}$  &  0.849(0.438) &  0.848(0.437) &  \textbf{0.848(0.437)} \\
\hline
\end{tabular}
\end{table}

According to Table \ref{tab:gdt}, the lowest average MSE score is found in MTGNN, with MTGNN$_{CORR}$ reaching a score of 0.838, without a clear distinction among the different sparsity levels. 
MTGNN is followed by ASTGCN$_{CORR}$ and
ASTGCN$_{DTW}$ with $0.84$ in case of GDT$=100\%$. This is remarkably different than utilizing sparser graphs, where MSE is around $0.89$. Similarly, A3tGCN$_{CORR}$ is also positively affected when using a denser graph, giving an MSE of $0.97$. However, the remaining graphs do not facilitate improving the performance of A3TGCN. 

As expected, the performance of the GNN models based on a random graph is the worst. It shows the biggest change for ASTGCN, moving to $1.06$ when using a random graph. In the case of MTGNN, by updating the random graph information, the error reaches the lowest $0.84$ score. On the contrary, the overall bad performance of A3TGCN is not getting much worse. Moreover, despite the overall high variability across all individuals shown in all methods, MTGNN and ASTGCN give the lowest standard deviation at $0.43$.  

Therefore, it is interesting to notice that the distance-based graph structures do not significantly affect the performance of GNN models. This can be caused by the fact that all these metrics can be effectively used in time-series data. The only case of positive change was seen in ASTGCN$_{CORR}$ and ASTGCN$_{DTW}$ when the whole graphs were used. On the other hand, using random graphs, meaning that these incorporate no useful information, leads to the worst performance.
This highlights the importance of selecting
an appropriate method for representing the relationships
between temporal sequences. 


\subsection{Experiment C}
Experiment C investigates the impact of the previously analyzed static (distance-based) graphs compared to the learned graph structures on the forecasting performance of multivariate
time series. This is designed by extracting the MTGNN-learned graph representations and using them as input into the other GNN models. The approach seems promising because of the successful application of MTGNN on EMA data that relies on iteratively updating the input graph. This comparison is
essential in understanding the efficacy of the learned graph approach derived by
the MTGNN model, against the initial static graphs.

\begin{figure*}
    \centering
\includegraphics[width = 0.75\textwidth]{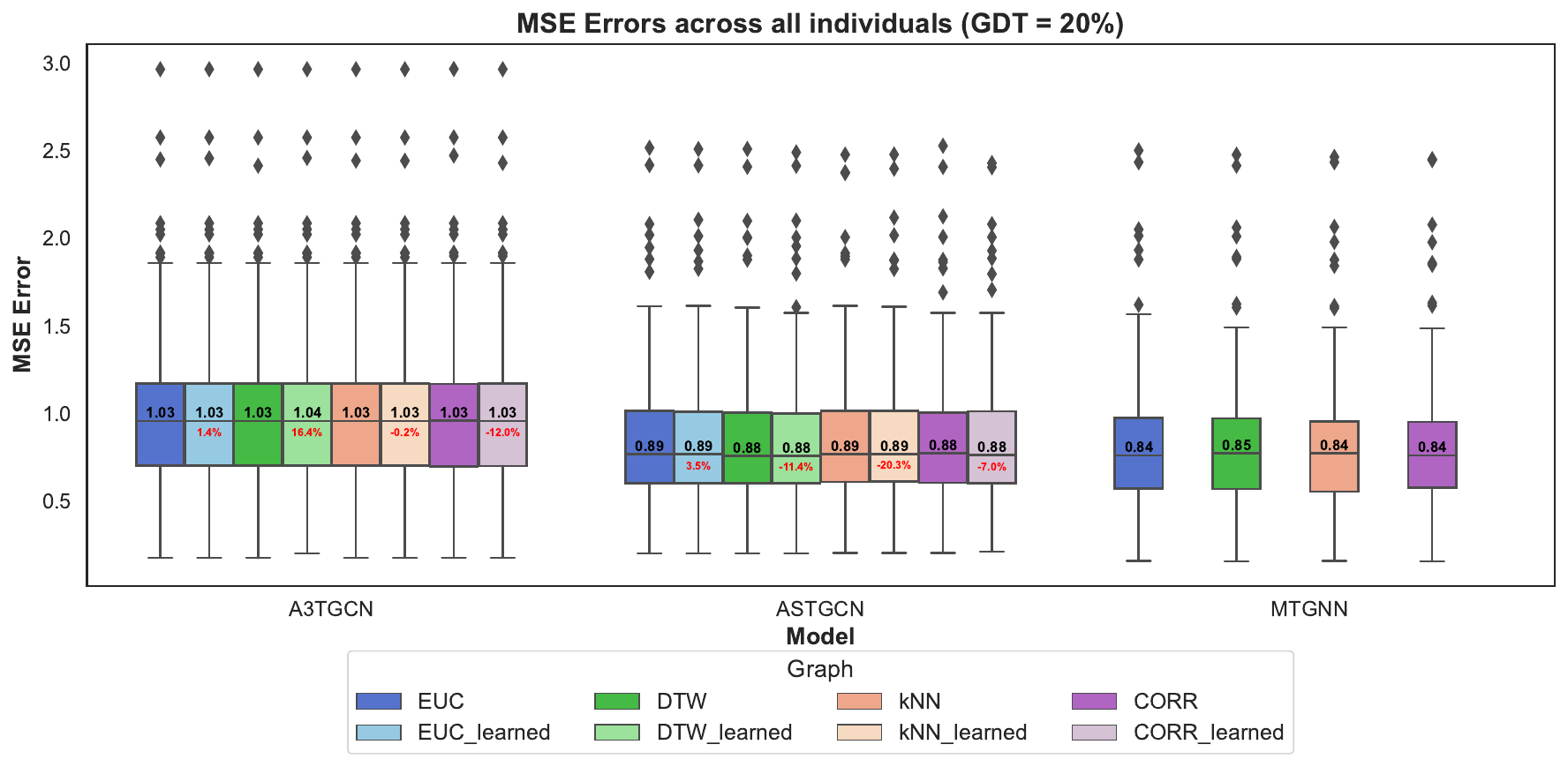}
    \caption{MSE distributions across all individuals comparing the graph learning process to the four
static graphs. Apart from the boxplot properties, mean values are given in black as well as the relative percentage of change in red.}
\label{fig:mse_learned}
\end{figure*}

In Fig.~\ref{fig:mse_learned}, the outcomes of comparing the graph learning process to the four
static ones are presented. In this case, the sparse version of the graph is utilized along with the 5-step input data. According to the extracted outcomes, MTGNN shows again the best performance, giving an MSE of 0.84. This is reasonable since the learned graph is optimized while trying to minimize the MTGNN training error. However, this result using the learned graph was not that effective. More specifically, 
ASTGCN shows only a slight error drop, without being affected by the learned graph structure. This is caused by the level of similarity between the two graphs, reaching $88\%$ correlation. 
Alternatively, the small differences between the two methods (A3TGCN and ASTGCN) can be further quantitatively assessed by the percentage change in their error rates. For each individual, the relative percentage of increase or decrease is calculated when using each distance metric and the corresponding MTGNN-learned graph. According to the results (marked in red) in Fig.~\ref{fig:mse_learned}, an overall decrease indicates an improvement in MSE error when using the MTGNN-learned graph.
Specifically, the biggest enhancement ($-20.3\%$) is observed in the case of ASTGCN, when using the kNN\_learned graph compared to kNN. It is also interesting to notice that a consistent improvement is revealed in error rates using DTW, kNN, and CORR, which contrasts with the results from the EUC metric.

Regarding A3TGCN, the performance remains the lowest, not exceeding an MSE of $1.02$. While the models show a slight drop in average MSE when using the learned version of a specified graph, it is worth exploring again the percentage of change in performance. More specifically, a performance enhancement is seen with kNN and CORR, when the learned graph is utilized instead of the pre-defined metrics. However, the increase in MSE percentage for DTW is unexpectedly high. Therefore, despite the general bad performance, individuals' MSE improvement is present in some models.


\section{Discussion}
In the exploration of applying GNNs to multivariate time-series forecasting, a set of three experiments was conducted whose key findings are summarized as follows. 

\subsection{GNN Performance}
In the first comparative analysis of GNNs, the study focuses on the performance of GNNs 
against the baseline LSTM model. According to Table \ref{tab:ExpA}, almost all GNN models provide better MSE scores than LSTM, which reaches the level of A3TGCN. GNNs exploit
the inherent relationships and dependencies among variables represented in a graph, allowing them to capture more complex patterns than LSTM. 

Subsequently, regarding GNNs, models from two different categories R-GCNs (A3TGCN) and T-GATs (ASTGCN, MTGNN) were investigated, applicable to sequentially-input MTS data. 
Among these, the models from the latter category show the best performance, with the MTGNN giving the lowest MSE at 0.84. The success of MTGNN is because it incorporates layers dedicated to graph learning, indicating the effectiveness of approaches updating the graph structure during training.

On the contrary, A3TGCN model produces higher MSE scores, also showing a greater sensitivity to a denser graph structure. This is attributed to the simplicity of its architecture relative to ASTGCN and MTGNN. A3TGCN integrates a temporal GCN with an attention layer, lacking processing any complex spatial-temporal dynamics inherent in the data.


Another aspect examined was the length of input data sequences. More specifically, during training, single- and multi-step sequences of input data were tested in terms of models' performance. After evaluation, when data is input as multi-step sequences, the errors slightly decrease with the errors showing similar trends across GNNs. Consequently, Experiments B and C are only focused on multi-step input sequences.
\subsection{Graph Construction}
Due to the dependence of GNN performance on the utilized graph, different characteristics of a graph structure were explored. First, this involved different distance-based graphs, since the ground truth representations across all variables of each individual were not known. While, afterwards, different aspects, such as sparsity and graph-learning structures, were investigated. 

According to all experiments' results, a consistent pattern was retrieved, where models based on dense correlation graphs outperformed all the others. Although all examined distance metrics are commonly used in time-series data, 
correlation was proved particularly powerful, resulting in a meaningful similarity matrix.
This distinction to the other graphs was mostly apparent for ASTGCN, whereas, for MTGNN, there was a minimal impact because of the learning process. 


Moreover, the effect of the level of graphs' sparsity on performance was evaluated. The findings of Experiment B showed the superiority of dense graphs in yielding better performance for ASTGCN and A3TGCN. 
However, for MTGNN, using a sparse graph facilitates focusing on the most valuable information, whereas a denser graph potentially introduces unnecessary complexity and noise, finally leading to similar performance.

Based on Experiment C, the significant role of learned graphs in improving the predictive performance of GNN models was further highlighted. MTGNN-learned graphs cause a small decrease in MSE compared to all initial static graphs, also, when used in other models. In cases of CORR and kNN, the overall percentage of MSE decrease indicates a performance improvement. 
However, it is highly dependent on the
dynamics of the time series data. This is evident in the variability of weight updates and subsequently, MSE performance, observed across different individuals. Only in
some cases, the updates to the weights were substantial, highlighting more significant relationships from the data.

\subsection{Limitations - Future Work}
As already discussed, enhancement in performance was observed in many cases of all the examined experiments, such as 
using the MTGNN-learned graphs. However, the differences were not significant. This is probably caused by the fact that the produced MSE scores are averaged over many dimensions, that are across all individuals, all variables, and all the time-points on the test set. Thus, potential effects at an individual level can be relatively suppressed by that. Particularly, using boxplots in Fig.~\ref{fig:mse_learned}, considerable variability in performance across individuals is observed. This can arise from the different quality of their data. For example, the number of time-points recorded per individual, the variance in variables, or the presence of noise in the data. Focusing on the importance of improving MSE scores at an individual level, this difference is also reflected by the relative percentage of MSE change. For instance, for ASTGCN$_{kNN}$, although the average MSE score was 0.89 when using both static and learned graphs, there was a $20.3\%$ MSE decrease when checking the individual improvement of the latter case. Similarly, the effects across the MSE scores when predicting each of the variables should be further investigated.

Furthermore, multi-step input data was proved to lead to better forecasting performance. This is reasonable because richer data representations are used, involving a subset of historical data. However, more experiments should be conducted on the most appropriate length of the input data sequence. 

Further exploration should also be done on the selected graph structure. Alternative types of distance metrics or threshold levels in sparsity should be carefully investigated as well as interpreted for their inter-variables connections. Finally, inspired by the success of MTGNN, it is also crucial to acknowledge that graph-learning mechanisms are incorporated in other GNN models as well. The graphs learned by advanced methods, such as Graph for Time Series (GTS), and Neural Relational Inference (NRI), should be further compared to both static and MTGNN-learned graphs \cite{nri, gts}.  
\section{Conclusions}
In conclusion, this work has significantly contributed to the understanding of
applying GNN models on EMA MTS data. 
More specifically, several GNN models were investigated aiming at successfully forecasting the progression of EMA data. After a set of experiments, GNN models proved to outperform the baseline LSTM model, decreasing the MSE from $1.01$ to $0.84$ when using MTGNN. Overall, MTGNN had the best performance, taking advantage of its internal graph learning layer that updates the initially input graph structure. Due to this success, the MTGNN-learned graph was also explored by the other GNN methods, showing that it could potentially enhance individuals' performance.
This highlights the significance of incorporating meaningful information regarding the inner relationships within the data, ultimately leading to improved EMA forecasting performance and ultimately a better understanding of mental disorders.

\bibliographystyle{IEEEtran}

\bibliography{\myreferences} 

\end{document}